\documentclass{article}
\usepackage[utf8]{inputenc} 
\usepackage[T1]{fontenc}    
\usepackage{lmodern}        
\usepackage{microtype}      
\usepackage{booktabs}       
\usepackage{hyperref}       
\usepackage{longtable}      
\usepackage{float}          
\usepackage{graphicx}       
\usepackage{amssymb}        
\usepackage{fancyvrb}       
\usepackage{url}            
\usepackage{xcolor}         
\usepackage[most]{tcolorbox} 
\usepackage{amsmath}        
\usepackage{authblk}        

\hypersetup{ 
    colorlinks=true,
    linkcolor=blue,
    filecolor=magenta,
    urlcolor=cyan,
}

\title{Muon Optimizer Accelerates Grokking}
\date{April 22\textsuperscript{th}, 2025} 

\author[1]{Amund Tveit\thanks{Corresponding author: \href{mailto:Amund.Tveit@microsoft.com}{\nolinkurl{Amund.Tveit@microsoft.com}}}}
\author[1]{Bjørn Remseth}
\author[1]{Arve Skogvold}
\affil[1]{Microsoft} 

\begin{document}

\maketitle

\section*{Abstract} 

This paper investigates the impact of different optimizers on the grokking phenomenon, where models exhibit delayed generalization. We conducted experiments across seven numerical tasks (primarily modular arithmetic) using a modern Transformer architecture. The experimental configuration systematically varied the optimizer (Muon vs. AdamW) and the softmax activation function (standard softmax, stablemax, and sparsemax) to assess their combined effect on learning dynamics. Our empirical evaluation reveals that the Muon optimizer, characterized by its use of spectral norm constraints and second-order information, significantly accelerates the onset of grokking compared to the widely used AdamW optimizer. Specifically, Muon reduced the mean grokking epoch from 153.09 to 102.89 across all configurations, a statistically significant difference (t = 5.0175, p = 6.33e-08). This suggests that the optimizer choice plays a crucial role in facilitating the transition from memorization to generalization.


\section{Introduction}

Grokking refers to the phenomenon where a model initially achieves high training accuracy while maintaining chance-level validation accuracy, before abruptly generalizing to high validation performance after extended training. Originally observed on small algorithmic tasks (e.g.,
modular arithmetic) \cite{prieto2022grokking}, grokking challenges our understanding of how
overparameterized models eventually discover generalizable patterns well
after overfitting.

Prior work has established that weight decay and other regularization techniques are crucial prerequisites for observing the grokking phenomenon \cite{liu2022towards, thiry2022grokking}. Our aim here is to investigate whether changing
the \emph{optimizer} alone - from the commonly used AdamW to the newer
Muon (\cite{jordan2024muon} and \cite{liu2025muon}) - can accelerate
grokking, i.e., reduce the number of epochs needed for that sudden
jump in validation accuracy.

We hypothesize that \textbf{Muon accelerates grokking} due to several
key differences in how it updates model weights compared to AdamW. See
details in figure \ref{whymuonhelps}.

\begin{figure}[!htbp]
\centering 
{\small
\begin{tabular}{|c|p{4cm}|p{4cm}|p{4cm}|}
\toprule
&{\bf Why Muon helps}& {\bf What Muon Does Differently} & {\bf How this helps with Grokking} \\ \midrule

1 & Promotes broader exploration & Uses orthogonalized gradient updates & Helps escape memorization and discover real patterns \\ \midrule
2 & Prevents runaway weights & Applies spectral norm constraints &  Keeps training stable and avoids ``softmax collapse'' \\ \midrule
3 & Keeps layers in sync & Matches update size to layer shape & ensures all layers learn at the right pace \\ \midrule
4 & Follows better directions & Approximates second-order updates & Reaches generalization faster with fewer steps \\ \midrule

5 & Trains more efficiently & Uses fewer FLOPs for same loss& Grokking can happen sooner with less compute \\
\bottomrule
\end{tabular}
}
\caption{Key mechanisms by which the Muon optimizer may accelerate grokking compared to AdamW.} 
\label{whymuonhelps}
\end{figure}

Empirical support also comes from NanoGPT, an informal benchmark suite for efficient GPT-2 training, where Muon was initially developed and demonstrated strong performance.

\section{Experimental Setup}\label{experimental-setup}

All experiments were conducted utilizing Nvidia H100 GPUs to ensure sufficient computational resources for the required training duration. The core model training and evaluation code was implemented in PyTorch \cite{tveit2024torchgrokking}. Individual experimental runs, each with a specific combination of softmax variant, optimizer, and dataset, were managed by a bash driver script. Results were logged to CSV files and subsequently processed using SQLite and Python for statistical analysis.

\subsection{Datasets \& Tasks}\label{datasets-tasks}

We utilize small algorithmic datasets, primarily focusing on modular arithmetic operations (addition, multiplication, division, exponentiation, greatest common divisor) and a parity task. These tasks are known to exhibit grokking under specific training conditions. For instance, in modular arithmetic (modulo 97), a task might be "A op B = C mod 97". The model must learn the underlying algebraic structure to generalize beyond the training examples. Each dataset comprises input pairs (e.g., 'A', 'B') and the corresponding output ('C'). The full dataset for modular tasks contains all $97 \times 97 = 9409$ possible input combinations. Training/validation splits were created by randomly permuting the full dataset and allocating a specific fraction to training (details in Figure \ref{datasetdimensions}). The parity task involves predicting the parity bit for 10-bit binary strings, with 1024 possible combinations. These specific algorithmic tasks were chosen because they reliably exhibit grokking under appropriate hyperparameter settings, facilitating its study \cite{nanda2023progress}.

\begin{figure}[!htb]
\centering 
{\small
\begin{tabular}{@{}lll@{}} 
\toprule
\textbf{Grokking Task} & \textbf{Dataset Description} & \textbf{Details} \\ \midrule
Gcd & Greatest common divisor, mod 97 & 9409 rows, 50\% train split \\
Mod-add & Addition, mod 97 & 9409 rows, 80\% train split \\
Mod-div & Division, mod 97 & 9409 rows, 80\% train split \\
Mod-exp & Exponentiation, mod 97 & 9409 rows, 70\% train split \\
Mod-mul & Multiplication, mod 97 & 9409 rows, 50\% train split \\
Parity & Parity of a 10-bit number & 1024 rows, 50\% train split \\ \bottomrule
\end{tabular}
}
\caption{Overview of datasets used in the experiments.} 
\label{datasetdimensions}
\end{figure}

\subsection{Softmax Configurations}\label{softmax-configurations}

While the primary focus is on the optimizer, we also explored the potential influence of the softmax activation function, motivated by research suggesting numerical stability issues in standard softmax can affect grokking \cite{power2025grokking}. We compared three variants (details in Figure \ref{softmaxvariants}):
\begin{itemize}
    \item \textbf{Softmax}: The standard exponential normalization.
    \item \textbf{Stablemax}: A variant using a piecewise transformation designed to enhance numerical stability \cite{power2025grokking}.
    \item \textbf{Sparsemax}: A variant that projects logits onto the probability simplex, often resulting in sparse outputs, which might offer implicit regularization benefits \cite{martins2016sparsemax}.
\end{itemize}

\begin{figure}[!htb]
\centering 
\begin{tabular}{|l|p{11cm}|} 
\toprule
\textbf{Softmax variant} & \textbf{Description and Formula} \\
\midrule
Softmax & Standard exponential normalization: Converts a real-valued vector $z$ into a probability distribution.
\[ \text{softmax}(z)_i = \frac {e^{z_{i}}}{\sum_{j} e^{z_{j}}} \] \\ \midrule
Stablemax \cite{power2025grokking} & Applies a piecewise transformation $s(z_i)$ for numerical stability before normalization:
\[ s(z_i) = \begin{cases} z_i + 1 & \text{if } z_i \geq 0, \\ \frac{1}{1 - z_i} & \text{if } z_i < 0. \end{cases} \quad \text{and} \quad {\text{stablemax}(z)}_i = \frac{s(z_i)}{\sum_j s(z_j)} \] \\ \midrule
Sparsemax \cite{martins2016sparsemax} & Projects $z$ onto the probability simplex, often yielding sparse outputs (some probabilities are exactly zero). It involves finding a threshold $\tau$ such that $\sum_i \max\{z_i - \tau, 0\} = 1$.
\[ {\text{sparsemax}(z)}_i = \max\left\{z_i - \tau, 0\right\} \]
The threshold $\tau$ is found efficiently using sorting and cumulative sums. \\
\bottomrule
\end{tabular}
\caption{Softmax variants compared in the experiments.} 
\label{softmaxvariants}
\end{figure}

\subsection{Model Architecture} 

The model used is a Transformer architecture implemented in PyTorch, adapted from an earlier MLX version \cite{stockeh2023mlxgrokking}. It incorporates several modern components: 
\begin{itemize}
    \item \textbf{Embedding Layer}: Maps input tokens (integers representing numbers or operators) to dense vectors. We use simple identity embeddings where the integer value itself is used as the embedding index (e.g., token '42' maps to embedding vector 42).
    \item \textbf{Transformer Blocks}: A stack of standard Transformer blocks.
    \item \textbf{Attention}: Multi-head self-attention with scaled dot-product mechanism.
    \item \textbf{Positional Encoding}: Rotary Positional Embeddings (RoPE) are used to inject sequence position information.
    \item \textbf{Normalization}: RMSNorm is used instead of LayerNorm for potentially better stability and performance.
    \item \textbf{Feed-Forward Network (FFN)}: Uses the SiLU (Sigmoid Linear Unit) activation function within the FFN sub-layers.
    \item \textbf{Regularization}: Residual connections are used throughout, and dropout is applied for regularization.
\end{itemize}
This architecture integrates established and modern Transformer components, providing a robust foundation for investigating grokking on the specified algorithmic tasks.

\begin{tcolorbox}[colback=blue!5!white,colframe=blue!75!black, title=Grokking Threshold Definition] 
Grokking is operationally defined as the first epoch where validation accuracy reaches or exceeds 95\%, occurring after the training accuracy has already stabilized near 100\%.
\end{tcolorbox}

\subsection{Optimizers Compared} 

The core comparison of this study is between two optimizers:
\begin{itemize}
    \item \textbf{AdamW}: A standard and widely used optimizer based on Adam, with decoupled weight decay. We used typical hyperparameters: $\beta_1 = 0.9$, $\beta_2=0.98$.
    \item \textbf{Muon}: A more recent optimizer designed to improve training dynamics by incorporating spectral norm constraints and approximations of second-order information \cite{jordan2024muon, liu2025muon}.
\end{itemize}
Both optimizers were configured with equivalent weight decay strengths to isolate the impact of their distinct update mechanisms. We performed multiple independent runs for each experimental condition (combination of task, optimizer, softmax variant) using different random seeds to ensure robustness of the findings. The primary metric recorded was the epoch number at which grokking (validation accuracy $\geq$ 95\%) occurred. Epoch distributions are visualized in Figure \ref{fig:epochdistributions}.

\section{Experimental Results}

\begin{figure}[!htb]
  \centering 
  \includegraphics[width=\textwidth,height=\textheight,keepaspectratio]{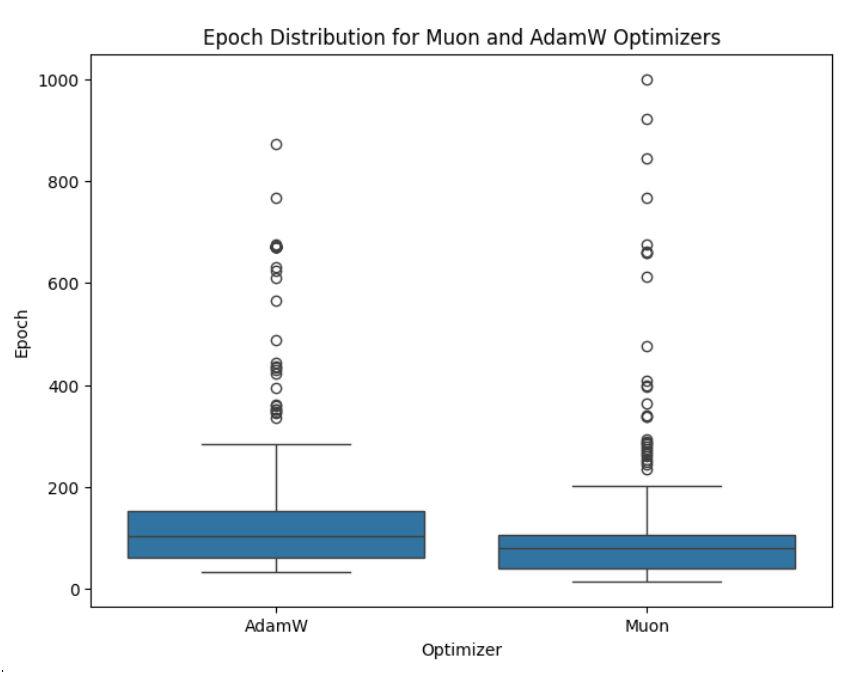} 
  \caption{Distribution of grokking epochs for Muon and AdamW optimizers across all tasks and softmax configurations. The boxplot shows medians, quartiles, and potential outliers, indicating Muon tends to grok earlier.} 
  \label{fig:epochdistributions}
\end{figure}

Our results demonstrate a clear and statistically significant advantage for the Muon optimizer in accelerating grokking. Across the combined experiments, Muon consistently led to earlier onset of high validation accuracy compared to AdamW. A two-sample t-test comparing the mean grokking epochs for the two optimizers yielded:
\begin{itemize}
    \item \textbf{Mean Grokking Epoch (AdamW)}: 153.09
    \item \textbf{Mean Grokking Epoch (Muon)}: 102.89
    \item \textbf{T-statistic}: $\approx 5.0175$
    \item \textbf{P-value}: $\approx 6.33 \times 10^{-8}$
\end{itemize}
The extremely low p-value indicates that the observed difference in means is highly statistically significant. Muon not only reduced the average number of epochs required to grok (see Figure \ref{meangrok}) but also exhibited a tighter distribution of grokking times, as visualized in Figure \ref{fig:epochdistributions}. Averaged across all experimental conditions, Muon consistently achieved the grokking threshold ($\geq$ 95\% validation accuracy) in fewer epochs than AdamW. 
Figure \ref{fig:epochdistributions} (boxplot) illustrates that the median epoch for Muon-based
grokking is notably lower than for AdamW (see figure \ref{meangrok}).

\begin{figure}[!htb]
\centering 
\begin{tabular}{@{}lc@{}} 
\toprule
{\bf Optimizer}& {\bf Mean Grokking Epoch} \\ \midrule 
AdamW & 153.09 \\
Muon & 102.89 \\ \bottomrule
\end{tabular}
\caption{Mean number of epochs required to reach the grokking threshold ($\ge$95\% validation accuracy) for each optimizer, averaged across all experimental conditions.} 
\label{meangrok}
\end{figure}

\section{Conclusion}

Our results strongly support the hypothesis that Muon significantly accelerates grokking compared to AdamW, demonstrating that the optimizer's update geometry critically influences the onset of delayed generalization. By incorporating spectral
norm constraints and second-order cues, Muon appears to steer the model
away from purely memorizing solutions, thereby facilitating earlier
discovery of the true pattern. Future research directions include investigating the generalizability of these findings to larger model architectures, diverse task domains, and the interplay between Muon and other regularization techniques.

\section*{References} 


\end{document}